\documentclass[10pt,twocolumn,letterpaper]{article}

\usepackage[pagenumbers]{cvpr} %
\usepackage{todonotes}
\newcommand{\ours}[0]{DeDoDe}

\usepackage[style=english]{csquotes}
\usepackage[detect-all,per-mode=symbol,group-digits=integer,group-minimum-digits=4,group-separator={,}]{siunitx}
\newcommand{\parsection}[1]{\vspace{0mm}\noindent\textbf{#1}~}

\usepackage[pagebackref,breaklinks,colorlinks]{hyperref}

\def\confName{CVPRW}
\def\confYear{2024}

\title{DeDoDe v2: Analyzing and Improving the DeDoDe Keypoint Detector}

\author{Johan Edstedt$^1$
\quad
Georg Bökman$^2$
\quad
 Zhenjun Zhao$^{3,4}$
 \\
{\normalsize $^1$Linköping University,
$^2$Chalmers University of Technology}\\
{\normalsize
$^3$The Chinese University of Hong Kong, $^4$Texas A\&M University}
}
\begin{document}
\twocolumn[{%
\centering
\renewcommand\twocolumn[1][]{#1}%
\maketitle
\includegraphics[width=.48\textwidth]{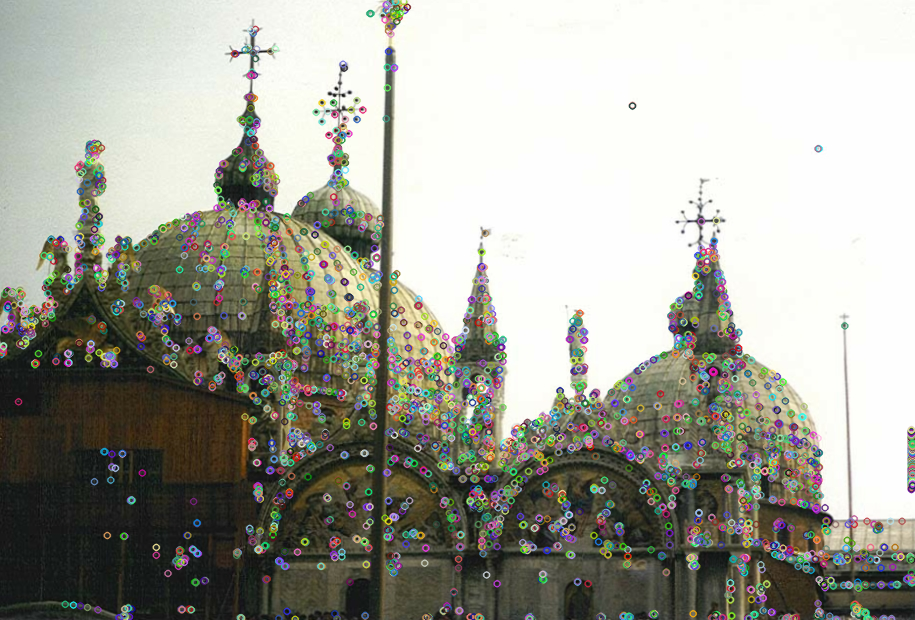}
\hfill
\includegraphics[width=.48\textwidth]{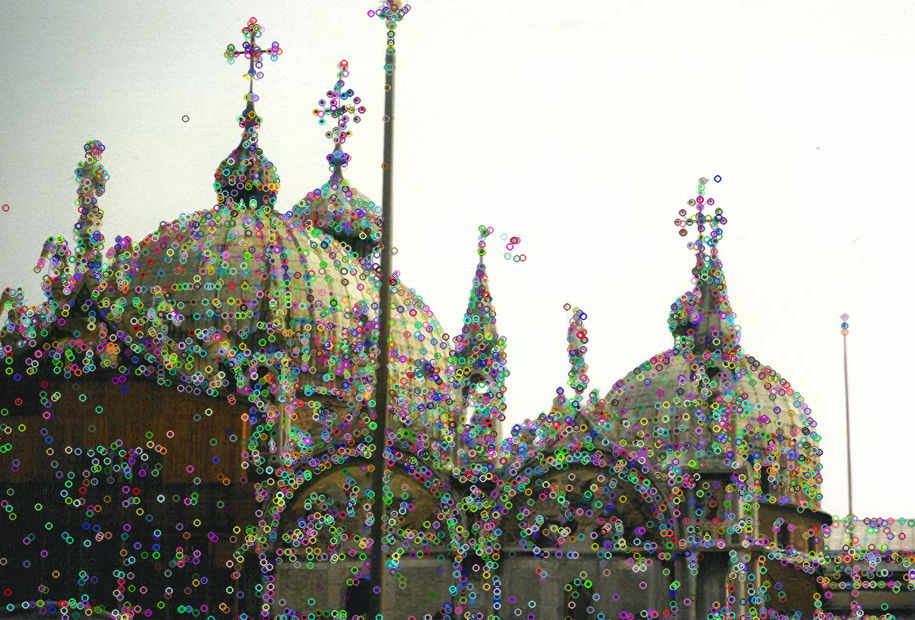}
\hfill
\includegraphics[width=.24\textwidth]{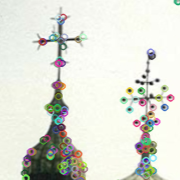}
\hfill
\includegraphics[width=.24\textwidth]{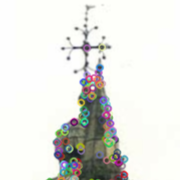}
\hfill
\includegraphics[width=.24\textwidth]{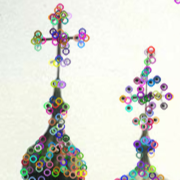}
\hfill
\includegraphics[width=.24\textwidth]{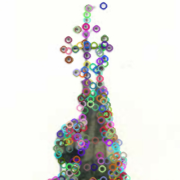}
\hfill

\captionof{figure}{\textbf{DeDoDe (left) vs DeDode v2 (right).} We propose DeDoDe v2, an improved keypoint detector following the \emph{detect don't describe} approach, whereby the detector is descriptor agnostic. We improve the DeDoDe detector, as demonstrated in the figure. DeDoDe struggles with \emph{clustering}, whereby keypoints are overly detected in distinct regions. This, in turn, causes it to underdetect in other regions, causing performance to degrade. In contrast, our proposed detector produces diverse but repeatable keypoints for the entire scene.\vspace{2em}}
\label{fig:teaser}}]

\begin{abstract}
\vspace{-1em}

In this paper, we analyze and improve into the recently proposed DeDoDe keypoint detector. We focus our analysis on some key issues. First, we find that DeDoDe keypoints tend to cluster together, which we fix by performing non-max suppression on the target distribution of the detector during training. Second, we address issues related to data augmentation. In particular, the DeDoDe detector is sensitive to large rotations. We fix this by including 90-degree rotations as well as horizontal flips. Finally, the decoupled nature of the DeDoDe detector makes evaluation of downstream usefulness problematic. We fix this by matching the keypoints with a pretrained dense matcher (RoMa) and evaluating two-view pose estimates. We find that the original long training is detrimental to performance, and therefore propose a much shorter training schedule. We integrate all these improvements into our proposed detector DeDoDe v2 and evaluate it with the original DeDoDe descriptor on the MegaDepth-1500 and IMC2022 benchmarks. Our proposed detector significantly increases pose estimation results, notably from 75.9 to 78.3 mAA on the IMC2022 challenge. Code and weights are available at \href{https://github.com/Parskatt/DeDoDe}{github.com/Parskatt/DeDoDe}.
\end{abstract}
    
\section{Introduction}
\label{sec:intro}

Obtaining corresponding pixels in multiple images of the same scene is a cornerstone task in computer vision, being an integral part of most structure-from-motion pipelines.
Classically, finding correspondences between two images is split into three sub-tasks: keypoint detection, description and matching.
This paper focuses on keypoint detection, building on the cutting-edge DeDoDe detector \cite{edstedt2024dedode}.
A keypoint is a local image feature distinct enough to be recognized under strong viewpoint and illumination variations.
However, it is difficult to precisely define into what this means (\eg, how strong variations should be allowed?), and we argue that the community still poorly understands what exactly a keypoint should be. As such, it is of interest to understand the training process of keypoint detectors and thoroughly investigate what components matter. In this paper, we conduct such an analysis, resulting in several simple modifications to the training pipeline of the DeDoDe detector,
significantly improving its performance. We hope our analysis will further encourage research into finding new and better objectives for keypoint detection.

Our main contributions are as follows.
\begin{enumerate}
    \item We introduce a series of training enhancements, including non-max suppression and improved data augmentation. See Section~\ref{sec:cluster}, Section~\ref{sec:minor}.
    \item We shorten the training time of the DeDoDe detector to 20 \emph{minutes} on a single A100 GPU while improving performance. See Section~\ref{sec:training}.
    \item We detail a multitude of tested modifications that did not work. See Section~\ref{sec:no-effect}.
\end{enumerate}

\section{Related Work}
The classical approach to finding correspondences between two images is to rely on a keypoint detector and a keypoint descriptor and match keypoint descriptions by some variant of nearest neighbours.
Influential works include the Harris and Shi-Tomasi corner detectors \cite{harris1988combined, shi1994good} as well as the SIFT detector and descriptor \cite{lowe2004distinctive} based on finding local extrema in scale space.
These classical methods have been improved and refined over the years, \eg through RootSIFT \cite{arandjelovic2012three}, HarrisZ \cite{bellavia2011improving} and HarrisZ+ \cite{bellavia2022harrisz+}.
Recently, however, deep learning approaches have become increasingly popular for both keypoint detection and description, \eg Tilde~\cite{verdie2015tilde}, SuperPoint ~\cite{detone2018superpoint}, AffNet~\cite{affnet-eccv-2018} and later works \cite{barroso2019key, dusmanu2019d2, revaud2019r2d2, tyszkiewicz2020disk, zhao2022alike, Zhao2023ALIKED}, demonstrating impressive performance boosts compared to the hand-crafted classical methods.
Furthermore, improving on the detector-descriptor pipeline, graph neural network-based descriptor matching such as SuperGlue and follow-up work~\cite{sarlin2020superglue, shi2022clustergnn, lindenberger2023lightglue} as well as
end-to-end semi-dense methods starting with LoFTR~\cite{sun2021loftr, wang2022matchformer, tang2022quadtree, bokman2022case, chen2022aspanformer} and dense image matchers like GLU-Net~\cite{truong2020glu,truong2021warp,truong2023pdc,edstedt2023dkm,edstedt2024roma} have started to see increasing use, however at non-negligible computational expense.
DeDoDe~\cite{edstedt2024dedode} showed that the more straightforward detector-descriptor pipeline remains competitive with the newer end-to-end approaches.
\citet{edstedt2024dedode} suggested decoupling the detector and descriptor training to reduce the reliance of the detector on a simultaneously trained descriptor.
Decoupling the two further enables research to focus on only the keypoint detector, which is the aim of this paper.
In the next section, we recap how the DeDoDe detector was trained in~\cite{edstedt2024dedode}.

\subsection{The DeDoDe Detector}
The main idea in DeDoDe is to train the detector with 3D tracks from structure-from-motion (SfM) reconstructions as a prior.
The idea is that keypoints from an available detector (in this case, SIFT) are filtered
by the SfM process to obtain \emph{good} keypoints.
To this end, the large MegaDepth~\cite{li2018megadepth} dataset of SfM reconstructions from internet images of tourist locations is used as the source for SfM tracks and hence keypoint priors.

In more detail, the training objective for
the DeDoDe detector consists of 1) the cross entropy
between a predicted keypoint probability map over the input image and a target probability map, and 2) a coverage regularization encouraging probability spread over the MVS from MegaDepth per image.
The following procedure generates the target probability map.
During training, pairs of overlapping views are sampled.
As the base target for the probability map of each view, the 2D projections of the 3D
tracks visible in both views are used.
This base target is smoothened by convolving with a Gaussian.
The probability maps of each view are then multiplied after being warped to the other view.
To add a degree of self-supervision to the training, the probability maps are multiplied by the
predicted probability map in each view.
Finally, the target probability maps are binarized by thresholding at an adaptive value, giving $k$ keypoints per batch.

\section{Analysis and Improvements}
\begin{figure}
    \centering
    \includegraphics[width=\linewidth]{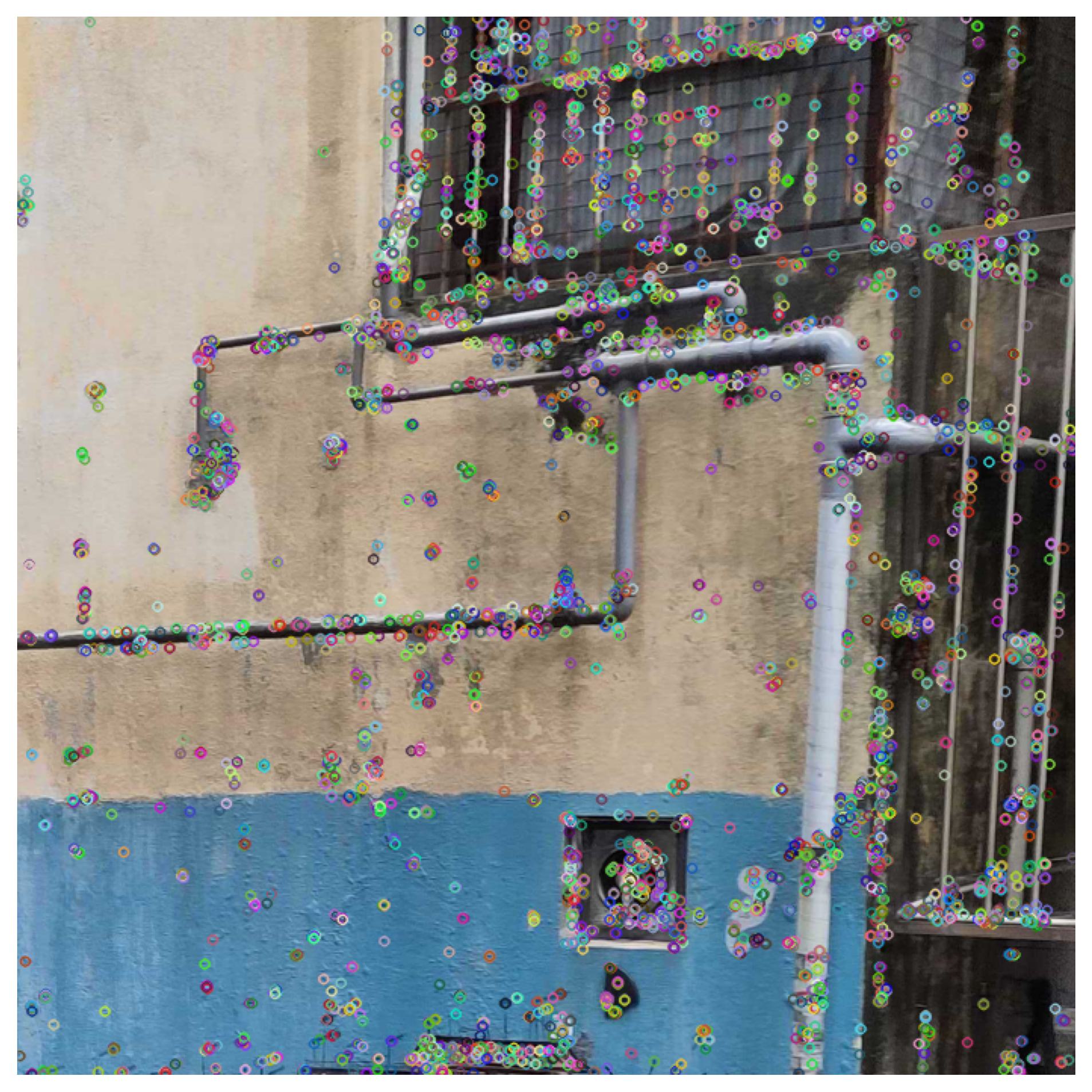}
    \caption{\textbf{Clusters in DeDoDe Detections.} The DeDoDe detection objective does not explicitly enforce sparsity in the detections. This has the side-effect of the network producing so-called \emph{clusters} of detections in particularly salient areas of the image. This is problematic in downstream tasks, as it means that many keypoints must be sampled to ensure repeatability. }
    \label{fig:clusters}
\end{figure}

\begin{figure}
    \centering
    \includegraphics[width=\linewidth]{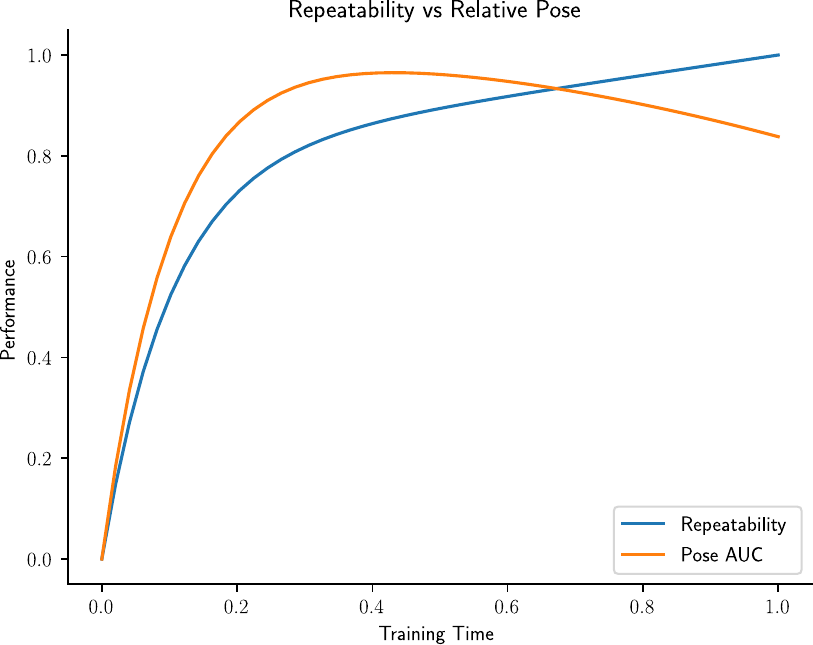}
    \caption{\textbf{Overfit to repeatability objective.} We qualitatively illustrate the tension between repeatability and downstream relative pose objectives. We found that during the course of training, while the keypoints tended to become more distinct and repeatable, this resulted in less distinct regions getting almost no keypoints, in particular outside regions with COLMAP MVS, resulting in worse relative pose estimates.}
    \label{fig:repeat-vs-auc}
\end{figure}

\begin{figure*}
    \centering
    \includegraphics[width=.45\linewidth]{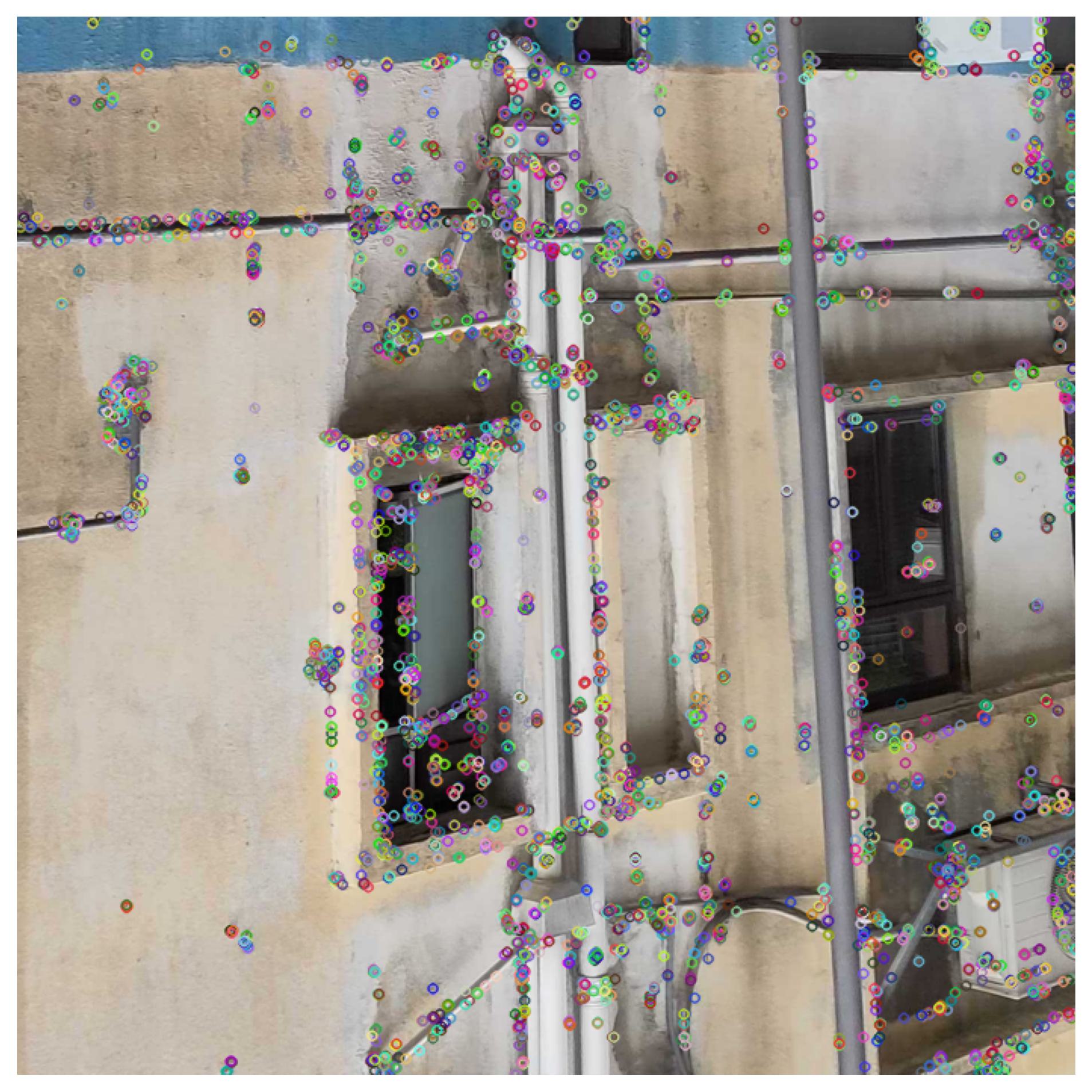}
    \includegraphics[width=.45\linewidth]{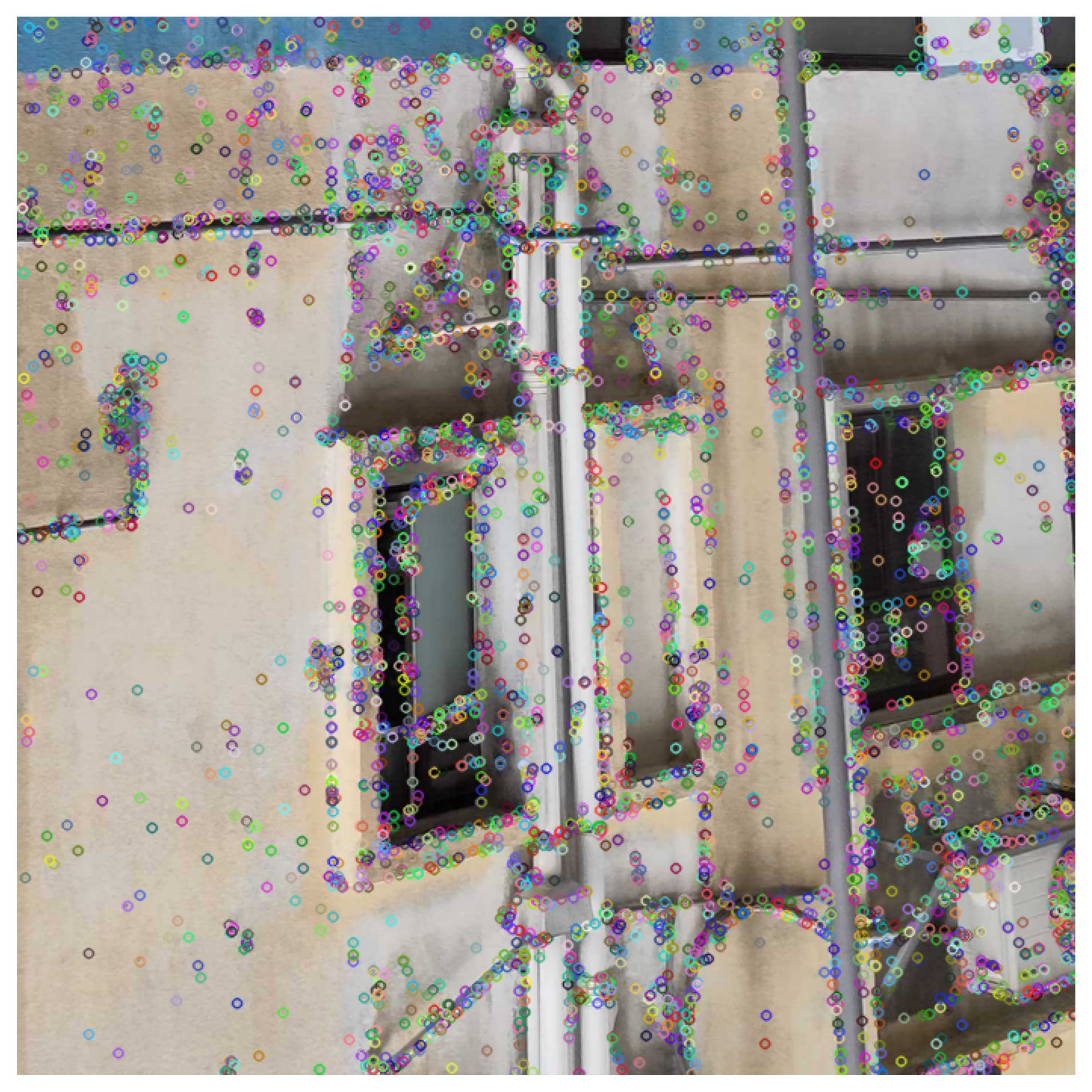}
    \includegraphics[width=.45\linewidth]{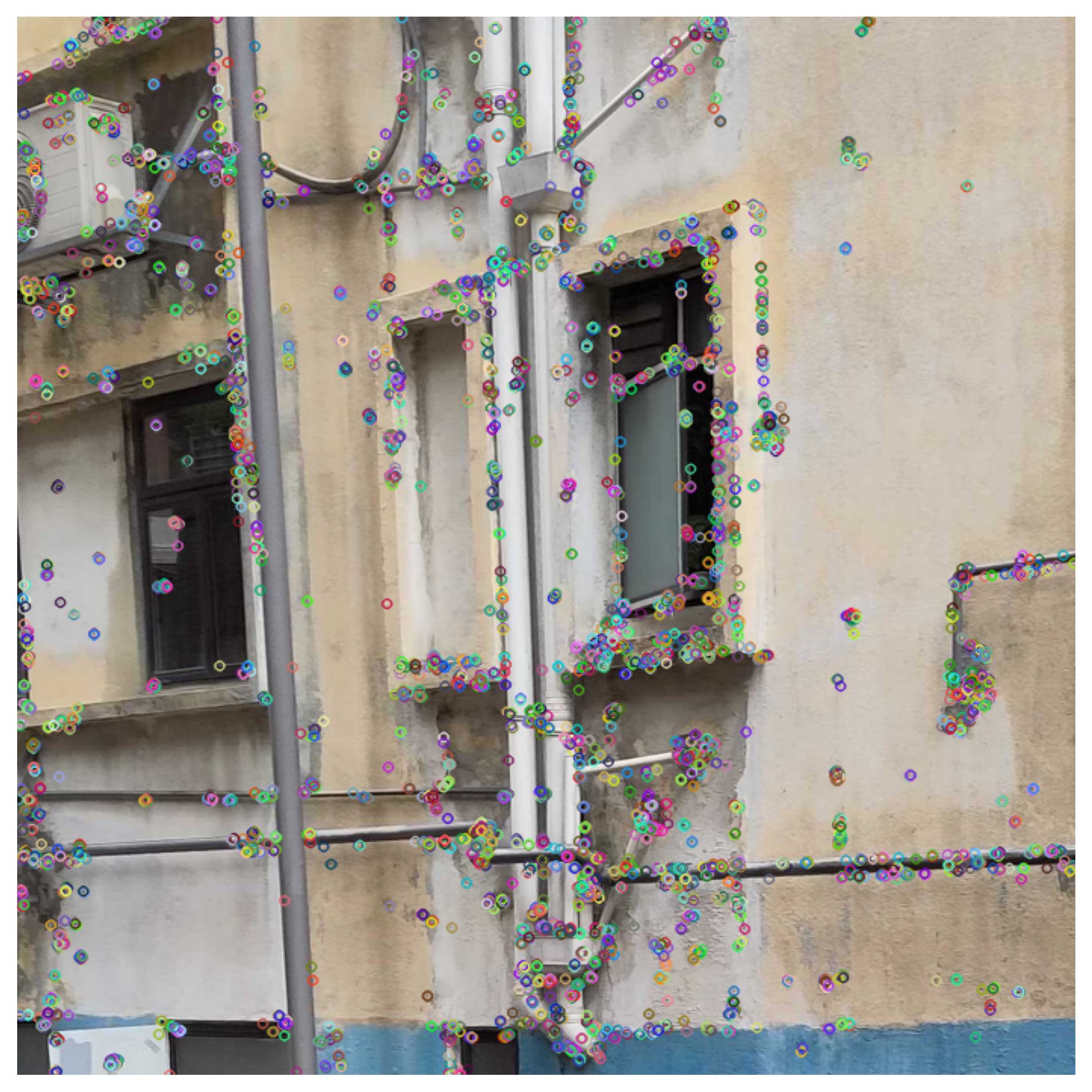}
    \includegraphics[width=.45\linewidth]{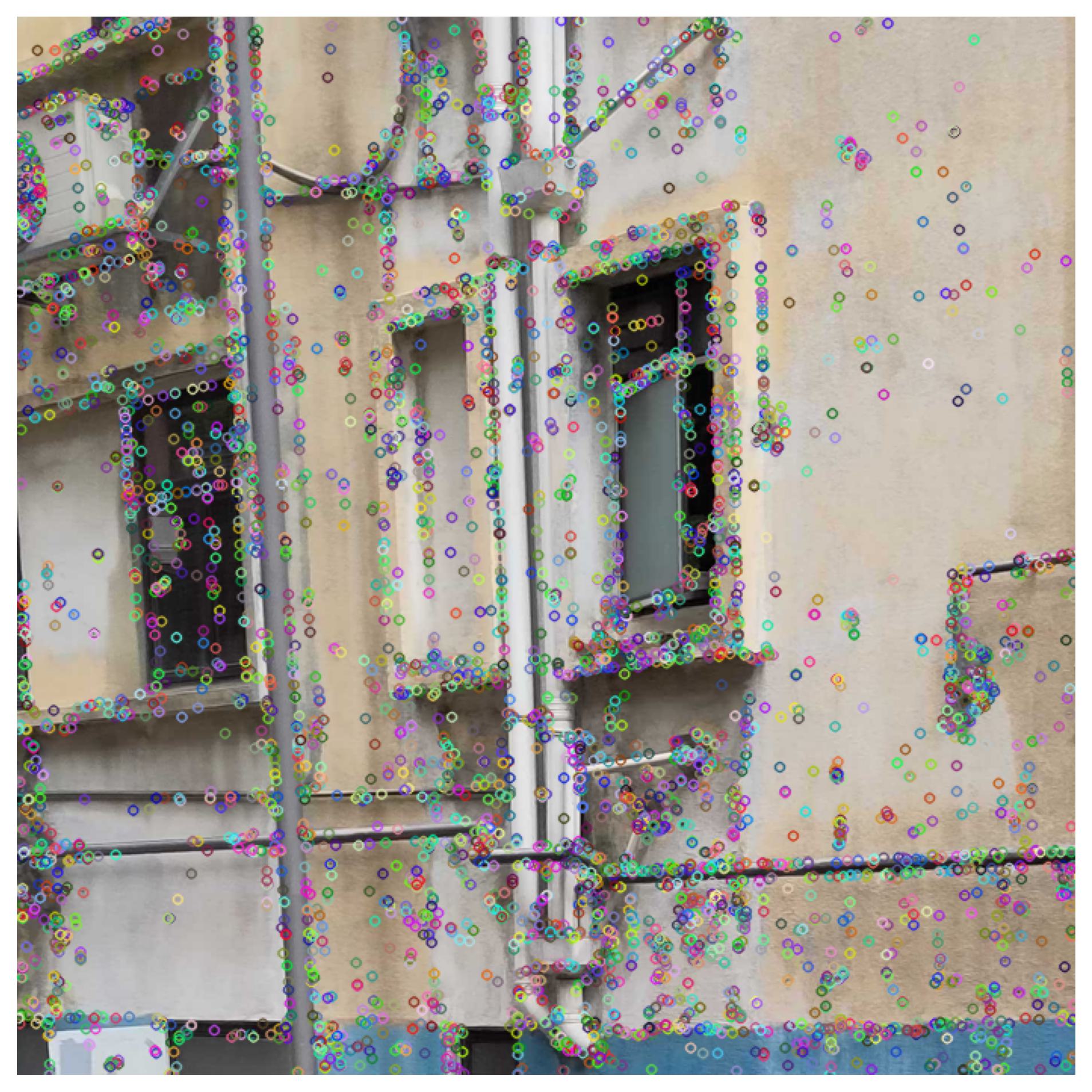}
    \caption{\textbf{Sensitivity to large rotations.} The original DeDoDe detector (left) is sensitive to large in-plane rotations. This was first noted by~\citet{bökman2024steerers}. We extend their ideas and additionally include joint horizontal flips. DeDoDe v2 produces more consistent keypoints under rotation of the input image (right). We plot the top \num{5000} keypoints in all images.}
    \label{fig:rots}
\end{figure*}
In this section, we analyze issues of the DeDoDe detector and propose a set of improvements to solve them.
\label{sec:detector}
\subsection{Preventing Clustering}
\label{sec:cluster}
We observe that the original DeDoDe detector tends to detect clusters (\emph{cf} Figure~\ref{fig:teaser}, Figure~\ref{fig:clusters}). This is undesired, as it reduces the diversity and coverage of the keypoints. However, we found that naively enforcing NMS during test time did not work well. 

\parsection{NMS During Training.} 
We are inspired by the peakiness loss from, \eg, R2D2~\cite{revaud2019r2d2} and the built-in soft-NMS in detectors such as ALIKED~\cite{Zhao2023ALIKED}, and propose a train-time NMS objective.  
To this end, we do top-$k$ after performing a $h\times h$ NMS on the posterior detection distribution. That is, after combining the detection prior with the logit predictions of the detector, we additionally enforce that the score be a local maximum to be set as a target. In practice, we set $h=3$, as it worked the best empirically. This simple change alleviated many issues in the original detector, \cf Figure~\ref{fig:teaser}.

\subsection{Training Time}
\label{sec:training}
The original DeDoDe detector is trained with \num{800000} image pairs on the MegaDepth dataset. We find that while the repeatability of the keypoints keeps increasing during the training, even on the test set, this does not transfer to the downstream objective of two-view relative pose estimation. We qualitatively illustrate this in Figure~\ref{fig:repeat-vs-auc}.

However, it is not entirely obvious how to measure this downstream objective as the detector is decoupled from the descriptor. Measuring the AUC would seemingly require recoupling the objectives, which is problematic~\cite{edstedt2024dedode}.

Instead, we propose using RoMa~\cite{edstedt2024roma} to match the keypoints to estimate downstream usability. When evaluating the original DeDoDe detector in this way, it quickly overfits the repeatability metric and that performance on pose estimation drops during training. To ensure that the detector did not overfit the scenes, we conducted an experiment where we included the test scenes in the training data. Perhaps surprisingly, we saw no major difference in performance and a similar downward trending curve over time. This indicates that there is a more fundamental issue between the detection objective and pose estimation, the investigation of which we leave for future study.

No matter the cause, we thus choose to drastically reduce the training time of the detector, setting it to \num{10000} image pairs, which significantly increases performance. Furthermore, the decrease in training time has the additional benefit of requiring significantly less compute to train, with training of DeDoDe v2 taking $\approx 20$ minutes on a single A100 GPU.

\subsection{Minor Improvements}
\label{sec:minor}
Here we discuss some minor changes and improvements we make to the training of the detector.

\parsection{Top-$k$ Computation.} DeDoDe computes the top-$k$ over a minibatch\footnote{This is not explicitly stated in the paper but can be observed in the released code: \url{github.com/Parskatt/DeDoDe}.} instead of per-pair. While this relaxes the assumption that each pair must contain a certain number of matching keypoints, it is problematic as difficult pairs may receive very few keypoints. We change this computation to be independent between the pairs.  

\parsection{Augmentation.} We follow the approach of Steerers~\cite{bökman2024steerers} and train the detector using random rotations in $\{0, 90, 180, 270\}$. We additionally include random horizontal flips. This makes the detector more robust to large rotations.

\subsection{Changes with no Effect}
\label{sec:no-effect}
Here, we describe a set of different hypotheses that turned out to have a negative or negligible effect on the detector's performance. We include these experiments for the curious reader.

\parsection{Diversity at Inference.} DeDoDe has a local density estimate post-hoc during inference. This is controlled by a parameter $\alpha$. In DeDoDe $\alpha = 1/2$. We experimented with setting it to other values $\in [0.5,1]$. We found no significant improvement in pose estimation results.

\parsection{Smoothness of Detection Prior.} The detection prior smoothing in DeDoDe is assumed to be Normal with standard deviation $\sigma = 0.5$. We experimented with setting it to other values (lower and higher). We found that both lower and higher values performed slightly worse.

\parsection{Annealing Prior Strength.} The prior strength in DeDoDe is set to $50$, which in practice means that the prior detections will always end up in the top-$k$ target. We found that decaying the strength over training significantly increases the repeatability of the keypoints. However, it simultaneously significantly decreases the downstream pose AUC. We hypothesize that this is due to the network disregarding less repeatable keypoints that nonetheless are important for accurate pose estimation.

\parsection{Changing $k$ in top-$k$.} We experimented with setting different $k\in [512,2048]$. We found that while there was a slight difference in performance, the setting of $k=1024$ from DeDoDe was seemingly optimal. 

\parsection{Changing Regularizer.} DeDoDe uses a coverage regularization:
\begin{equation}
    \mathcal{L}_{\rm coverage} = {\rm CE}(\mathcal{N}(0,\sigma^2) * p_{f_{\theta}}, \mathcal{N}(0,\sigma^2) * p_{\rm MVS}).
\end{equation}
with $\sigma = 12.5$ pixels. We investigated changing $\sigma$ as well as removing the regularization, as well as replacing it with a uniform regularizer. We found that these changes had either negligible or negative effects.

\parsection{Learning Rate.} The default learning rate in DeDoDe is $10^{-4}$ for the decoder and $2\cdot 10^{-5}$ for the encoder. Changing these had a negligible effect on performance.

\parsection{Training Resolution.} Since DeDoDe is trained on $512\times 512$ resolution and tested on $784\times 784$ resolution, we believed that using a random crop strategy during training (where the crops come from $784\times 784$ images) would alleviate potential train-test resolution gaps. This, however, turned out to have little effect on performance.

\parsection{NMS During Inference.} We found that even when trained with NMS, our detector still produces worse detections when NMS is applied post-hoc. However, the reduction in performance is significantly lower than for the baseline model. 

\section{Experiments}
\label{sec:experiments}
We train the detector for \num{10000} image pairs on the MegaDepth dataset, using the same training split as DeDoDe~\cite{edstedt2024dedode}. We use a fixed image size of $512\times 512$. We use a batch size of $7$. Training is done on a single A100 GPU and takes $\approx 20$ minutes.
\begin{figure*}
    \centering
    \includegraphics[width=.33\linewidth]{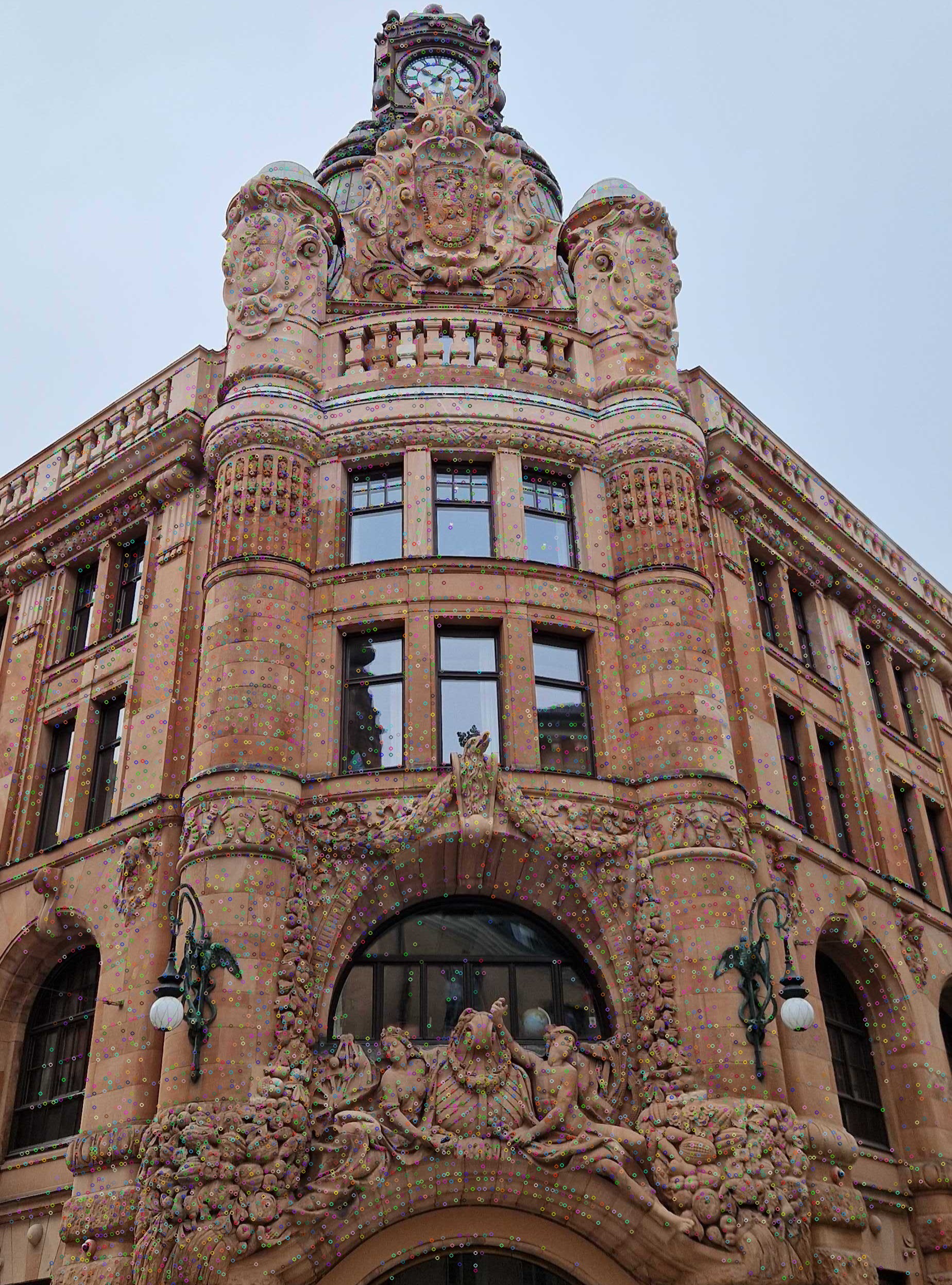}
    \includegraphics[width=.33\linewidth]{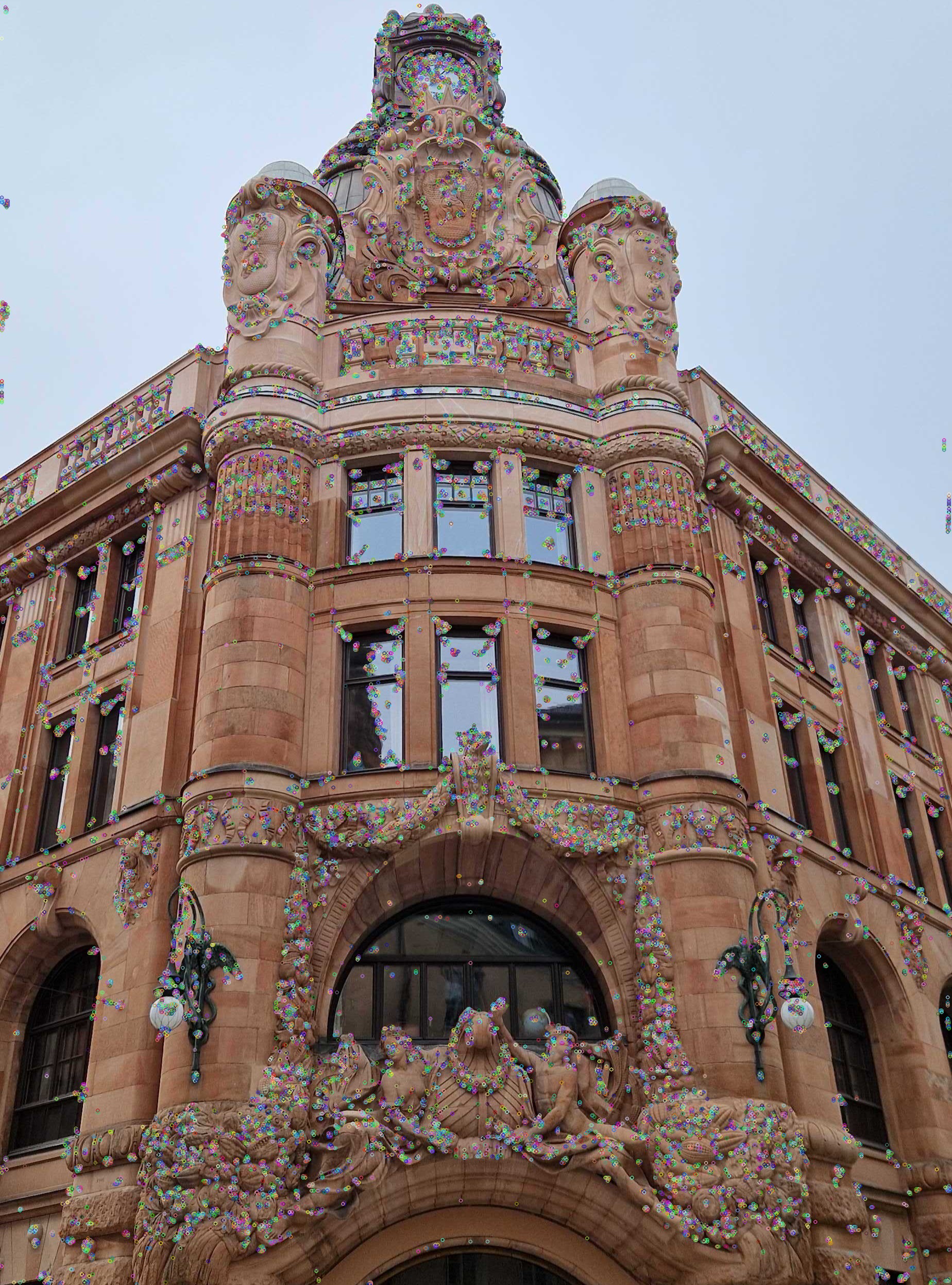}
    \includegraphics[width=.33\linewidth]{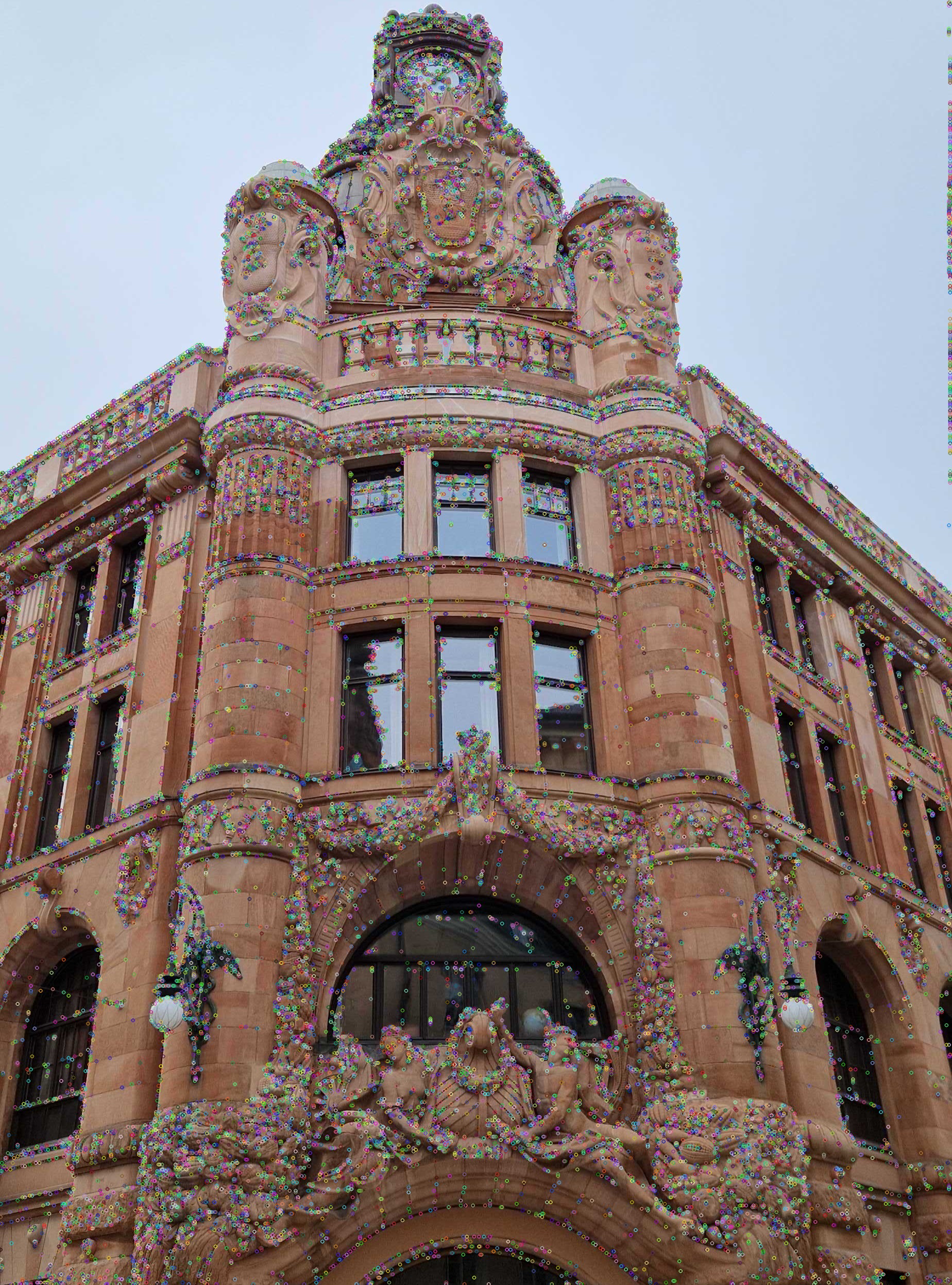}
    \includegraphics[width=.33\linewidth]{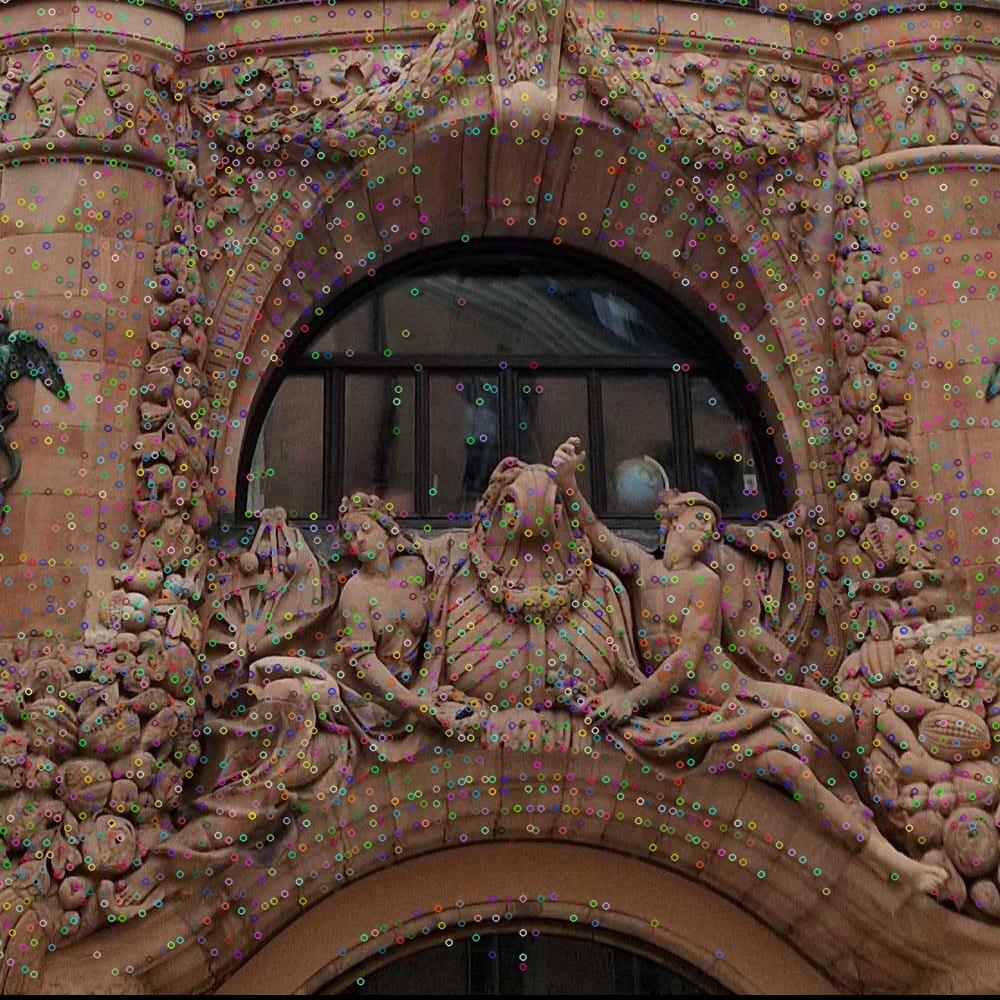}
    \includegraphics[width=.33\linewidth]{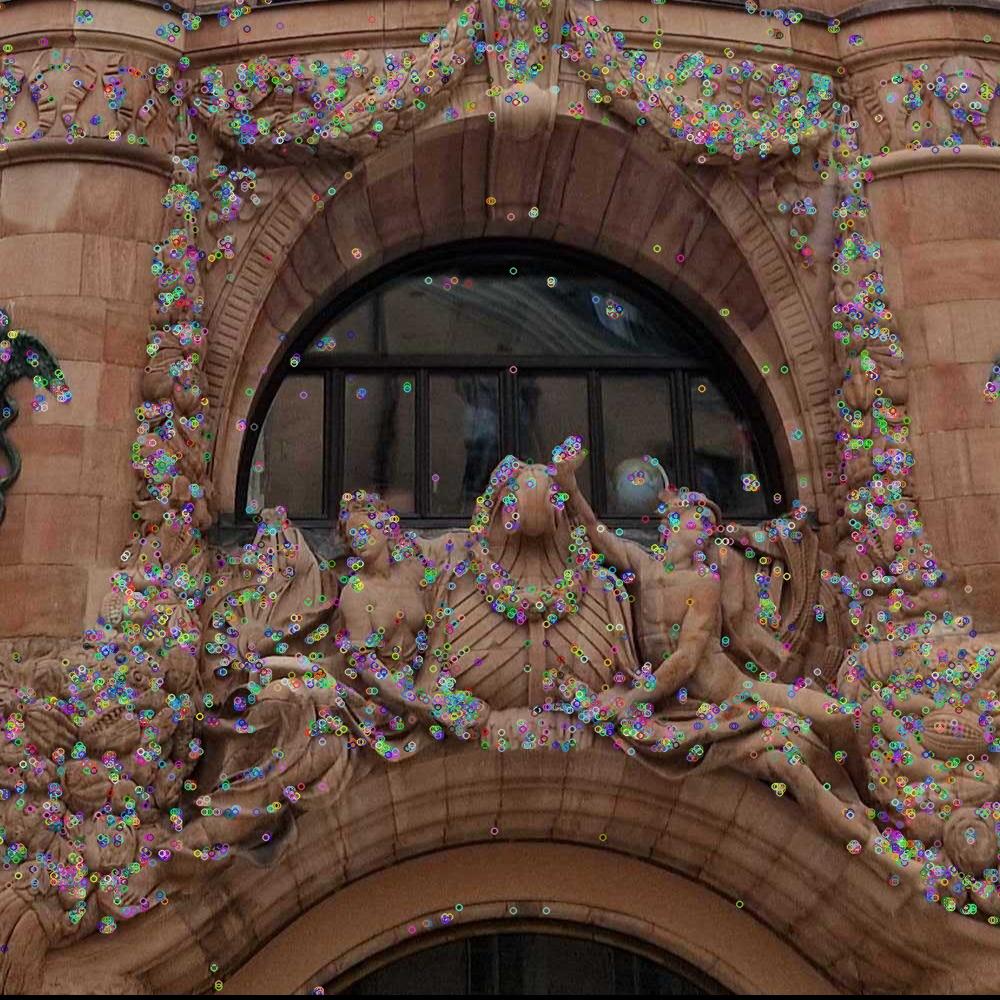}
    \includegraphics[width=.33\linewidth]{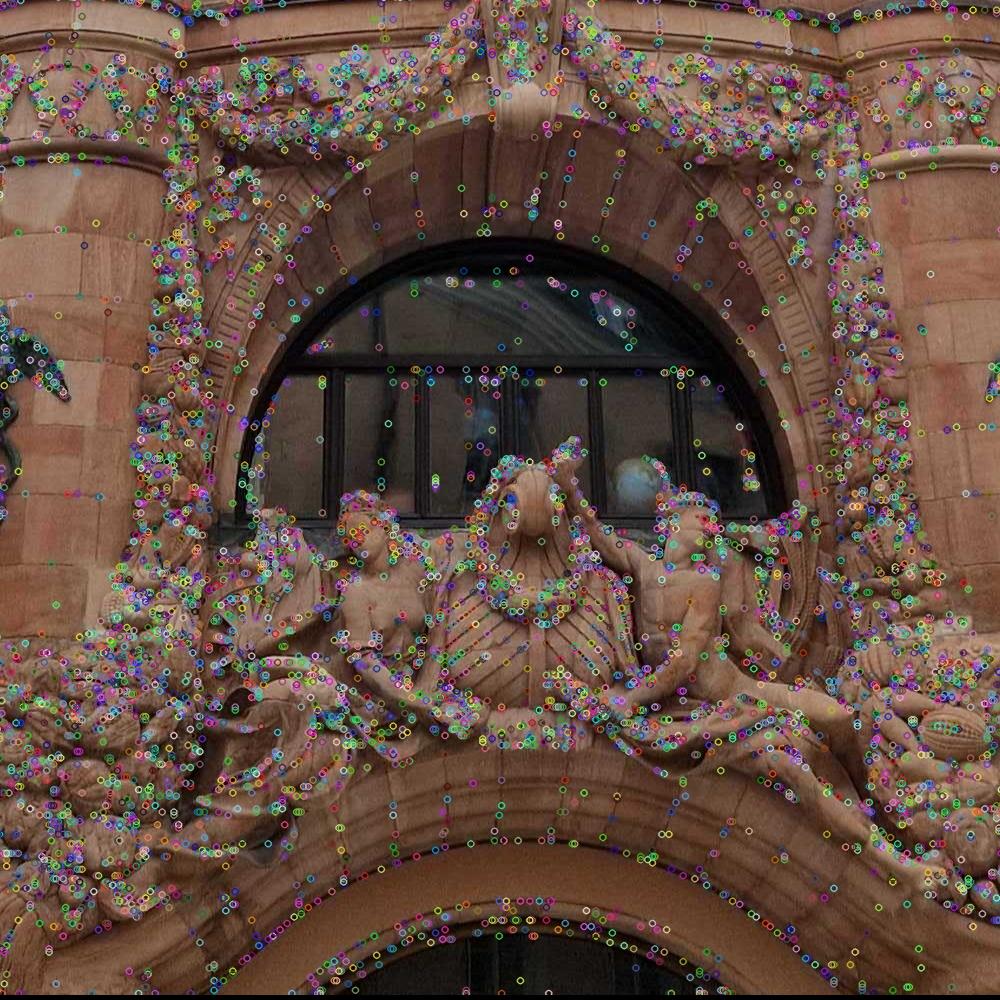}

    \caption{\textbf{Qualitative comparison of DISK~\cite{tyszkiewicz2020disk} (left), DeDoDe~\cite{edstedt2024dedode} (middle), DeDoDe v2 (right).} Best viewed in high resolution. DISK (left) produces diverse, but non-discriminative keypoints. DeDoDe, in contrast, produces discriminative kepoints, but tends to cluster. Our proposed DeDoDe v2 has the benefit of both approaches, yielding both diverse and discriminative keypoints.}
    \label{fig:kps}
\end{figure*}

\begin{figure*}
    \centering
    \includegraphics[width=.75\linewidth]{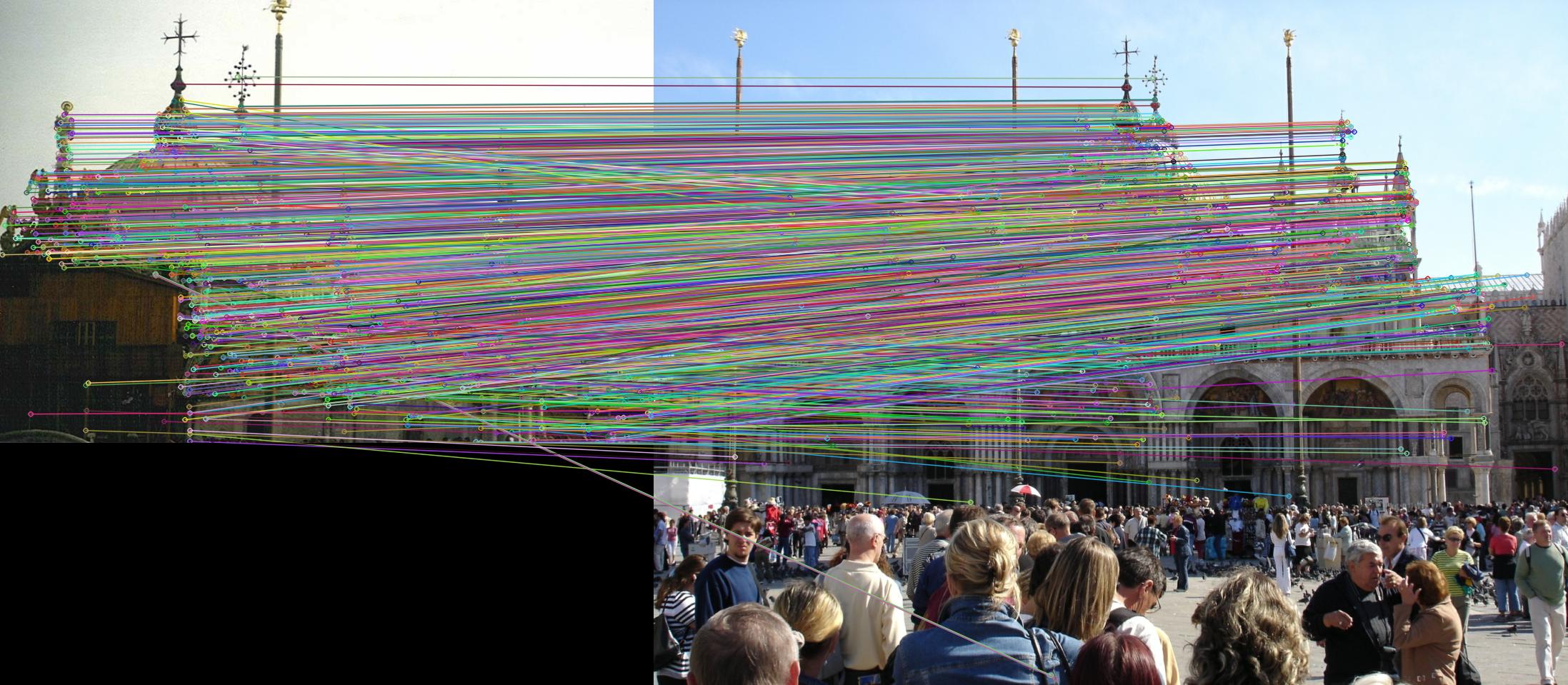}
    \includegraphics[width=.75\linewidth]{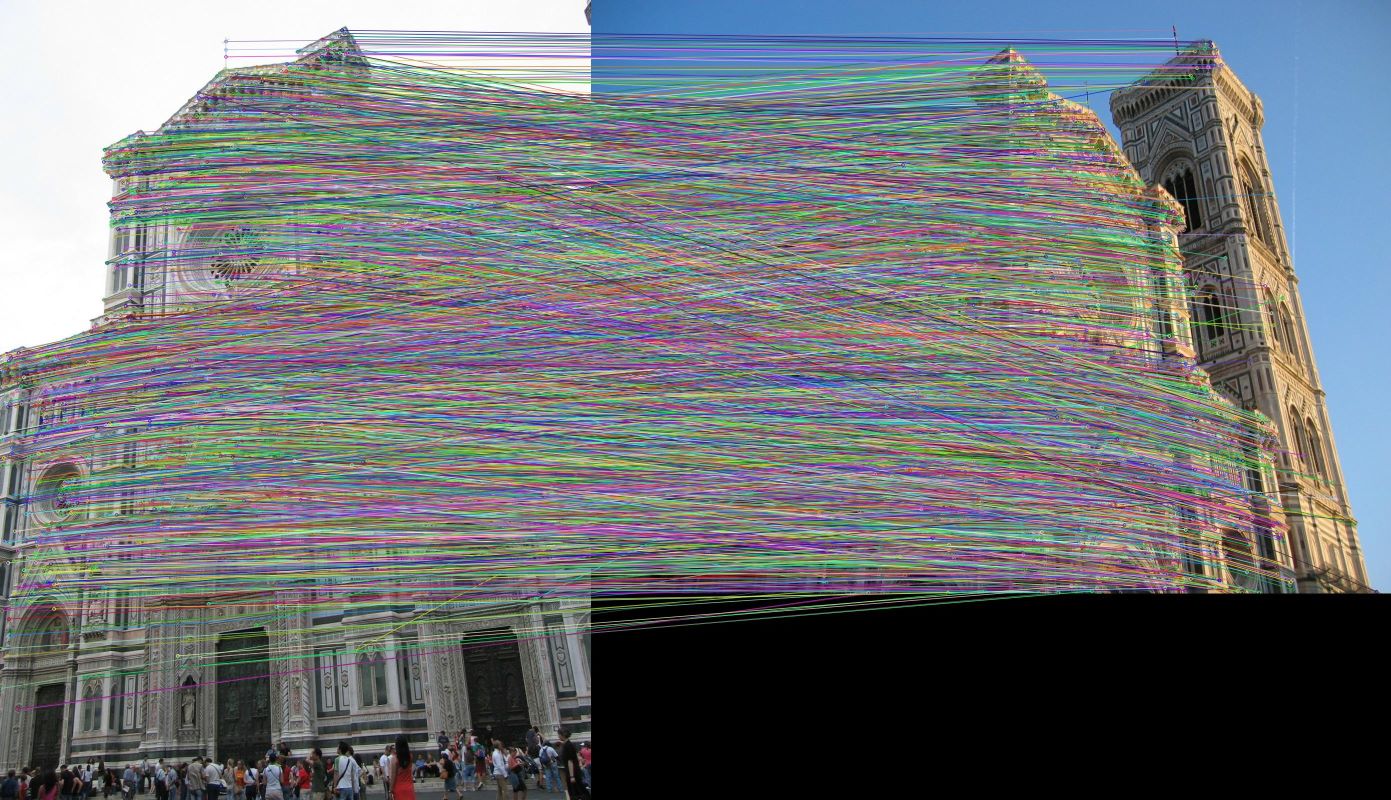}
    \includegraphics[width=.75\linewidth]{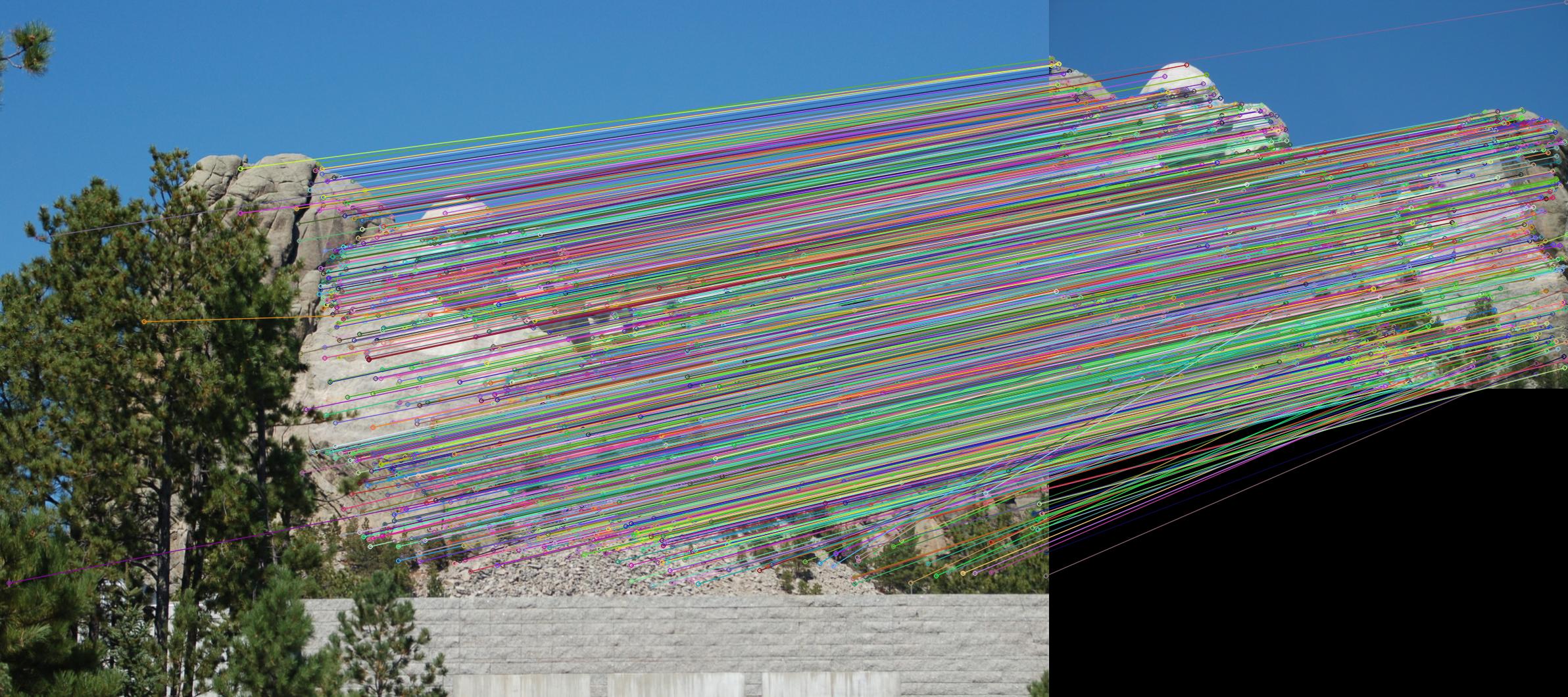}
    \caption{\textbf{Qualitative example of DeDoDe v2 matches.} }
    \label{fig:matches}
\end{figure*}

We run all SotA experiments at a resolution of $784\times 784$ and sample top-$k$ keypoints. We use the descriptors from~\cite{edstedt2024dedode}.

\subsection{SotA Comparison}
\parsection{MegaDepth-1500 Relative Pose:}
 MegaDepth-1500 is a relative pose benchmark proposed in LoFTR~\cite{sun2021loftr} and consists of 1500 pairs of images in two scenes of the MegaDepth dataset, which are non-overlapping with our training set. We mainly compare our approach to DeDode and previous detector descriptor methods. As in DeDoDe, we tune the methods for the preferred number of keypoints, and let both SiLK~\cite{gleize2023silk} and DeDoDe~\cite{edstedt2024dedode} detect up to \num{30000} keypoints, as we find that they benefit from it. We call this \emph{unlimited}, as any number of keypoints can be used.
We present results in Table~\ref{tab:megadepth-loftr}. We show clear gains on all performed benchmarks. We additionally evaluate our proposed detector with the RoMa matcher using \num{8000} keypoints with the original DeDoDe detector in Table~\ref{tab:megadepth-loftr-8k}. Again, we observe a clear boost in performance.
 \begin{table}
 \small
     \centering
     \caption{\textbf{SotA comparison on the Megadepth-1500-Unlimited benchmark}. We follow DeDoDe~\cite{edstedt2024dedode} and evaluate each detector with its optimal number of keypoints. In the case of SiLK and DeDoDe, we cap the number to 30k. Measured in AUC (higher is better).}
     \begin{tabular}{l rrr}
     \toprule
      Method $\downarrow$\quad\quad\quad AUC $\rightarrow$& $@5^{\circ}$&$@10^{\circ}$&$@20^{\circ}$\\

\midrule
SuperPoint~\cite{detone2018superpoint}~\tiny{CVPRW'18} & 31.7 & 46.8 & 60.1 \\
DISK~\cite{tyszkiewicz2020disk}~\tiny{NeurIps'20} & 36.7 & 52.9 & 65.9 \\
ALIKED~\cite{Zhao2023ALIKED}~\tiny{TIM'23} & 41.9 & 58.4 & 71.7 \\
SiLK~\cite{gleize2023silk}~\tiny{ICCV'23} & 39.9 & 55.1 & 66.9 \\
\midrule
DeDoDe v1-L --- v1-B~\cite{edstedt2024dedode}~\tiny{3DV'24} & 49.4 & 65.5 & 77.7\\
DeDoDe \textbf{v2}-L --- v1-B & \textbf{52.5} & \textbf{67.4} & \textbf{78.7} \\
\midrule
DeDoDe C4-L --- C4-B~\cite{bökman2024steerers}~\tiny{CVPR'24} & 51\phantom{.0} & 67\phantom{.0} & 79\phantom{.0} \\
DeDoDe \textbf{v2}-L --- C4-B & \textbf{52.6} &  \textbf{67.9} & \textbf{79.5} \\

\midrule
DeDoDe v1-L --- v1-G~\cite{edstedt2024dedode}~\tiny{3DV'24} & 52.8 & 69.7 & 82.0\\
DeDoDe \textbf{v2}-L --- v1-G & \textbf{54.6}  & \textbf{70.7} & \textbf{82.4} \\
      \midrule
          DeDoDe v1-L --- RoMa~\cite{edstedt2024roma}~\tiny{CVPR'24} &  55.1 & 71.6 & 83.5\\
          DeDoDe \textbf{v2}-L --- RoMa & \textbf{57.6} & \textbf{73.3} & \textbf{84.4}\\
     \bottomrule
     \end{tabular}

     \label{tab:megadepth-loftr}
 \end{table}
 
\begin{table}
 \small
     \centering
     \caption{\textbf{SotA comparison on the Megadepth-1500-8k benchmark}. We investigate the effect of reducing the number of keypoints when using the SotA RoMa matcher. Measured in AUC (higher is better).}
     \begin{tabular}{l rrr}
     \toprule
      Method $\downarrow$\quad\quad\quad AUC $\rightarrow$& $@5^{\circ}$&$@10^{\circ}$&$@20^{\circ}$\\
      \midrule
          DeDoDe v1-L --- RoMa &52.9   & 69.9 & 82.2\\
          DeDoDe \textbf{v2}-L --- RoMa & \textbf{54.9} & \textbf{71.4} & \textbf{83.1} \\

     \bottomrule
     \end{tabular}

     \label{tab:megadepth-loftr-8k}
 \end{table}

\parsection{Image Matching Challenge 2022:}
 The Image Matching Challenge 2022~\cite{image-matching-challenge-2022} comprises challenging uncalibrated relative pose estimation pairs. Different from MegaDepth-1500, the test set is hidden and does not derive from MegaDepth, and may, therefore, better indicate generalization performance, especially for models such as \ours~that train on MegaDepth.
 We follow the setup in SiLK~\cite{gleize2023silk} and DeDoDe~\cite{edstedt2024dedode} and use \num{30000} keypoints, and MAGSAC++~\cite{barath2020magsac++} with a threshold of $0.2$ pixels. We use a fixed image size of $784\times 784$. We follow the approach in RoMa~\cite{edstedt2024roma} and report results on the hidden test set. We present results in Table~\ref{tab:imc2022}. We improve by $+2.5$ mAA compared to the DeDoDe baseline.
\begin{table}
\small
    \centering
    \caption{\textbf{SotA comparison on the IMC2022 benchmark.} Relative pose estimation results on the IMC2022~\cite{image-matching-challenge-2022} hidden test set, measured in mAA (higher is better).}    \label{tab:imc2022}
    \begin{tabular}{l r}
    \toprule
     Method $\downarrow$\quad\quad\quad mAA $\rightarrow$&$@10$\\
     \midrule

DISK~\cite{tyszkiewicz2020disk}~\tiny{Neurips'20} & 64.8 \\
ALIKED~\cite{Zhao2023ALIKED}~\tiny{IEEE-TIM'23} & 64.9 \\
SiLK~\cite{gleize2023silk}~\tiny{ICCV'23} & 68.5 \\
\midrule
DeDoDe v1-L --- v1-B~\cite{edstedt2024dedode}~\tiny{3DV'24} & 72.9 \\
DeDoDe v2-L --- v1-B &  \textbf{74.7}\\
\midrule
DeDoDe v1-L --- v1-G~\cite{edstedt2024dedode}~\tiny{3DV'24} & 75.8 \\
DeDoDe \textbf{v2}-L --- v1-G & \textbf{78.3} \\
\bottomrule
    \end{tabular}
\end{table}
\subsection{Qualitative Examples}
We provide qualitative examples of DeDoDe v2 keypoints in Figure~\ref{fig:kps} and matches in Figure~\ref{fig:matches}. As can be seen in the figures, our proposed detector produced diverse but repeatable keypoints (Figure~\ref{fig:kps}), that are matchable (Figure~\ref{fig:matches}).

\section{Conclusion}
We have presented DeDoDe v2, an improved keypoint detector. We analyzed issues with the original detector and proposed several improvements to solve these. Our proposed detector sets a new state-of-the-art on the challenging IMC2022 and MegaDepth-1500 relative pose estimation benchmarks.

\parsection{Limitations.} We have empirically analyzed and improved the DeDoDe detector. However, we still lack a theoretical understanding of some of the underlying issues. In particular, formulating an objective that is not in tension with relative pose (\cf Figure~\ref{fig:repeat-vs-auc}) is of future interest.

\parsection{Acknowledgements.} 
We thank the reviewers for the constructive feedback.
This work was supported by the Wallenberg Artificial
Intelligence, Autonomous Systems and Software Program
(WASP), funded by the Knut and Alice Wallenberg Foundation and by the strategic research environment ELLIIT funded by the Swedish government. The computational resources were provided by the
National Academic Infrastructure for Supercomputing in
Sweden (NAISS) at C3SE, partially funded by the Swedish Research
Council through grant agreement no.~2022-06725, and by
the Berzelius resource, provided by the Knut and Alice Wallenberg Foundation at the National Supercomputer Centre.

{
    \small
    \bibliographystyle{ieeenat_fullname}
    \bibliography{main}
}

\end{document}